\documentclass[sigconf, screen]{acmart}
\AtBeginDocument{%
  }

\copyrightyear{2025}
\acmYear{2025}
\setcopyright{acmlicensed}
\acmConference[KDD '25]{Proceedings of the 31st ACM SIGKDD Conference on Knowledge Discovery and Data Mining V.2}{August 3--7, 2025}{Toronto, ON, Canada}
\acmBooktitle{Proceedings of the 31st ACM SIGKDD Conference on Knowledge Discovery and Data Mining V.2 (KDD '25), August 3--7, 2025, Toronto, ON, Canada}
\acmDOI{10.1145/3711896.3737163}
\acmISBN{979-8-4007-1454-2/2025/08}

\usepackage{enumitem}
\usepackage{multirow}
\usepackage{threeparttable}
\usepackage{tabularx}
\usepackage{subfigure}
\usepackage{graphicx}
\usepackage{array}

\begin{document}

\title[Towards Interpretable Drug-Drug Interaction Prediction]{Towards Interpretable Drug-Drug Interaction Prediction: A Graph-Based Approach with Molecular and Network-Level Explanations}

\author{Mengjie Chen}
\affiliation{%
  \institution{Data Science Institute, School of Mathematics, Shandong University}
  \city{Jinan}
  \state{Shandong}
  \country{China}}
\email{mengjiechen@mail.sdu.edu.cn}

\author{Ming Zhang}
\authornote{Corresponding author.}
\affiliation{%
  \institution{Academy of Mathematics and Systems Science, Chinese Academy of Sciences}
  \state{Beijing}
  \country{China}}
\email{mingzhang@amss.ac.cn}

\author{Cunquan Qu}
\authornotemark[1]
\affiliation{%
  \institution{Data Science Institute, Shandong University}
  \city{Jinan}
  \state{Shandong}
  \country{China}}
\email{cqqu@sdu.edu.cn}

\renewcommand{\shortauthors}{Mengjie Chen, Ming Zhang, \& Cunquan Qu}

\begin{abstract}
  Drug-drug interactions (DDIs) represent a critical challenge in pharmacology, often leading to adverse drug reactions with significant implications for patient safety and healthcare outcomes. While graph-based methods have achieved strong predictive performance, most approaches treat drug pairs independently, overlooking the complex, context-dependent interactions unique to drug pairs. Additionally, these models struggle to integrate biological interaction networks and molecular-level structures to provide meaningful mechanistic insights. In this study, we propose \textbf{MolecBioNet}, a novel graph-based framework that integrates molecular and biomedical knowledge for robust and interpretable DDI prediction. By modeling drug pairs as unified entities, MolecBioNet captures both macro-level biological interactions and micro-level molecular influences, offering a comprehensive perspective on DDIs. The framework extracts local subgraphs from biomedical knowledge graphs and constructs hierarchical interaction graphs from molecular representations, leveraging classical graph neural network methods to learn multi-scale representations of drug pairs. To enhance accuracy and interpretability, MolecBioNet introduces two domain-specific pooling strategies: context-aware subgraph pooling (CASPool), which emphasizes biologically relevant entities, and attention-guided influence pooling (AGIPool), which prioritizes influential molecular substructures. The framework further employs mutual information minimization regularization to enhance information diversity during embedding fusion. Experimental results demonstrate that MolecBioNet outperforms state-of-the-art methods in DDI prediction, while ablation studies and embedding visualizations further validate the advantages of unified drug pair modeling and multi-scale knowledge integration. Moreover, the pooling strategies yield interpretable insights into the underlying interaction mechanisms, highlighting MolecBioNet’s potential as a robust and transparent framework for predictive modeling in support of drug safety assessment.
\end{abstract}




\maketitle

\section{Introduction}

Combination therapy, which involves the concurrent use of two or more drugs, has increasingly emerged as a promising strategy for treating complex diseases such as hypertension and cancer \cite{OT16723, Cheng2019}. While this approach enhances treatment outcomes, it also elevates the risk of DDIs \cite{Jia2009, Tatonetti2012}. These interactions can alter the pharmacokinetics or pharmacodynamics of co-administered drugs, potentially leading to adverse drug reactions (ADRs) and, in severe cases, drug withdrawals \cite{Cascorbi2012, Vilar2014}. Given their clinical significance and associated risks, accurately predicting DDIs is crucial for optimizing treatment regimens, ensuring clinical drug safety, and protecting patient health \cite{Wang2015}.

Traditional methods for detecting DDIs, such as clinical trials and in vitro assays, though reliable, are time-consuming, labor-intensive, and limited in scalability for screening large numbers of potential interactions \cite{Whitebread2005}. Computational approaches have emerged as low-cost, efficient alternatives to address these limitations, offering significant potential to accelerate DDI prediction \cite{Ferdousi2017, Kastrin2018, Yu201804}. Early computational methods primarily relied on drug fingerprints \cite{Vilar2012} or manually engineered features \cite{Deng202008} to infer interactions from drug properties. However, such feature-centric methods often struggled to fully capture the intricate relations underlying DDIs, limiting their prediction accuracy.

Recent advancements in graph-based modeling techniques, particularly in complex networks and graph neural networks (GNNs), have enabled the development of powerful and flexible frameworks that leverage graph-structured data for DDI prediction \cite{He202206, Guoijcai2023}. Graph-structured data in the context of DDI prediction generally falls into two categories: homogeneous and heterogeneous graphs. Homogeneous graphs, such as traditional DDI networks, model drugs as nodes and their interactions as edges, providing a focused yet simplified view of drug relations \cite{Yu201804, du2024customized}. Despite their computational efficiency, these representations often fail to capture the complexity of biological interactions. On the other hand, heterogeneous graphs offer a richer representation by integrating diverse biological entities—such as drugs, proteins, and diseases—as nodes, and their various interactions (e.g., drug-protein, protein-disease) as edges, as exemplified by biomedical knowledge graphs. Additionally, molecular graphs represent the internal structure of drugs at the atomic level, where nodes correspond to atoms and edges denote chemical bonds, providing essential molecular insights into DDIs. Together, approaches—biomedical knowledge graph-based modeling  \cite{Tanvir2021, Zitnik2018, Karim2019, KGNNijcai2020, Yu202109, DDKGSu202205, LaGATHong202212} and molecular graph-based modeling \cite{Nyamabo202111, Xu201907, Nyamabo202201, Yu202404, Li202301}—have driven remarkable progress in DDI prediction.

Despite the success and potential of graph-based models, several challenges remain in DDI prediction. The first challenge lies in representing drug pairs holistically, capturing their intrinsic properties and interactions as a unified entity rather than treating each drug independently. Existing methods typically focus on encoding the features of each drug separately, either at the network level or the molecular level, and then combine these representations, often through concatenation or other heuristic operations \cite{ijcai2021p487, Chen202109, Su202403}. Such approaches inherently fail to reflect the complex, context-dependent relationships that arise uniquely within the drug pair. Consequently, they may overlook key information at both the macroscopic biological and microscopic molecular levels, leading to incomplete or suboptimal drug-pair representations and a limited ability to model the complexities of DDIs fully.

A second major challenge involves novel drugs—those in the early stages of development, with novel molecular structures, or rarely observed in existing DDI databases. The scarcity of interaction data for such drugs poses significant challenges for methods reliant on historical DDI information, particularly those grounded in DDI networks \cite{Yu201804, Al-Rabeah2022}. Approaches based on molecular features, such as drug fingerprints or graph-based representations of molecular structures, could be directly applied under cold-start conditions \cite{Dewulf2021, Liu202202}. However, these methods tend to lack the expressiveness necessary to fully capture the nuanced interactions between drugs and their broader biological context. Specifically, they often overlook inter-drug interactions and the connections between drugs and other biological entities, such as proteins or diseases. Biomedical knowledge graphs offer a promising source of side information to address these challenges \cite{Zhang202312}, yet existing models predominantly focus on macroscopic biological interactions, often at the expense of capturing molecular-level intra- and inter-drug interactions. This limitation contributes to an incomplete representation of drug pairs and their complex behaviors.

The third challenge is the interpretability of DDI prediction models. Beyond accurate predictions, interpretability is crucial for understanding the underlying mechanisms driving drug interactions. An interpretable model builds trust in its outputs and provides biologically meaningful insights for domain experts, bridging the gap between computational predictions and real-world applications \cite{VO20222112}. However, current methods \cite{Wang202403, Zhong202409} rarely integrate explanations across macroscopic biological interaction relations and microscopic molecular interaction mechanisms to elucidate the causes of drug interactions, limiting their capacity to explain the causes of DDIs comprehensively.

To address these challenges, we propose MolecBioNet, a graph-based framework for DDI prediction that integrates molecular and biomedical knowledge. By treating drug pairs as unified entities, MolecBioNet captures both macro-level interactions within biological networks and micro-level influences between molecular structures. Specifically, it extracts local subgraphs from biomedical knowledge graphs and constructs hierarchical interaction graphs based on molecular structures. Using classical graph neural network techniques, it learns drug-pair relationships at both biological and molecular levels. To enhance predictive accuracy and explainability, MolecBioNet introduces CASPool to identify biologically relevant entities and AGIPool to prioritize molecular substructures that most influence interactions. As a result, the framework facilitates robust and explainable DDI predictions, offering a valuable tool for DDI risk evaluation and mechanistic interpretation.

The primary contributions of this work are as follows:
\begin{itemize}[leftmargin=3.34mm]
    \item \textbf{Unified Drug Pair Modeling:} MolecBioNet shifts the focus from analyzing drugs in isolation to treating drug pairs as integrated entities. This paradigm enables the model to holistically capture mutual biological and molecular relationships, offering richer insights into DDI mechanisms.
    \item \textbf{Multi-Scale Knowledge Integration:} MolecBioNet combines macro-level biological context and micro-level molecular information by extracting local subgraphs from knowledge graphs and constructing hierarchical interaction graphs. Through mutual information minimization regularization, the framework maximizes the diversity and complementarity of information captured during fusing multi-scale embeddings.
    \item \textbf{Interpretable and Robust DDI Predictions:} MolecBioNet integrates specialized pooling mechanisms at multiple biological scales to enhance both interpretability and robustness. CASPool highlights contextually relevant biomedical entities within local subgraphs, while AGIPool prioritizes molecular substructures that are crucial for driving drug interactions. This multi-scale approach not only improves predictive performance but also provides mechanistic insights into drug interactions, making the model both a powerful predictive tool and an explanatory resource for biomedical research.
\end{itemize}

\section{Preliminaries}

\subsection{Problem Formulation}

This study aims to predict the type of interaction that may occur when two drugs are co-administered. To achieve this, we frame the problem as a multi-class classification task, aiming to develop a model that predicts the specific interaction type associated with a given drug pair.

Formally, let $\mathcal{D}$ denote a set of drugs and $\mathcal{R}$ a set of possible interaction types. Given a dataset $\mathcal{S} = \{(u, r, v) \mid u, v \in \mathcal{D}, r \in \mathcal{R} \}$, where each tuple $(u, r, v)$ represents a drug pair $(u, v)$ and its corresponding interaction type $r$. Our goal is to learn a mapping function $f: \mathcal{D} \times \mathcal{D} \rightarrow \mathcal{R}$ that predicts the interaction type $r$ for any given drug pair $(u, v)$.

\subsection{Task-Specific Biomedical Knowledge Graph Construction}

Accurate prediction of DDIs requires a comprehensive contextual understanding that extends beyond isolated interaction data. To this end, we construct a \textit{task-specific Biomedical Knowledge Graph} by integrating a \textit{Drug-Drug Interaction graph} with an \textit{external Knowledge Graph}. This integration not only combines direct interaction data but also enriches it with diverse biomedical relationships, enabling a holistic modeling of drug interactions and their underlying biological mechanisms. The components of this task-specific graph are defined as follows.

The \textit{Drug-Drug Interaction (DDI) graph}, denoted as $\mathcal{G}_{DDI} = (\mathcal{V}_{DDI}, \mathcal{E}_{DDI}, \mathcal{R}_{DDI})$, forms the foundational component of our graph construction. Here, $\mathcal{V}_{DDI} = \mathcal{D}$ represents the set of drugs, $\mathcal{R}_{DDI} = \mathcal{R}$ is the set of interaction types, and $\mathcal{E}_{DDI} = \{(u, r, v) \mid u, v \in \mathcal{D}, r \in \mathcal{R} \}$ defines the observed interactions. Each edge $(u, r, v)$ represents a specific interaction type $r$ between the drug pair $(u, v)$. While this graph directly reflects drug interaction data, it does not capture the broader interactions that drugs might have with other biological entities.

To provide a richer context, we integrate an \textit{external Knowledge Graph (eKG)}, represented as $\mathcal{G}_{KG} = (\mathcal{V}_{KG}, \mathcal{E}_{KG}, \mathcal{R}_{KG})$. In this graph, $\mathcal{V}_{KG}$ denotes the set of biomedical entities, including drugs, proteins, diseases, and others, while $\mathcal{R}_{KG}$ represents various types of relations, such as drug-protein interactions and drug-disease associations. Each edge $(u, r, v) \in \mathcal{E}_{KG}$ indicates the presence of relation $r$ between entities $u, v \in \mathcal{V}_{KG}$. The eKG offers a rich contextual representation, extending the DDI graph by providing knowledge on diverse relationships between drugs and other biological entities.

The \textit{task-specific Biomedical Knowledge Graph (tsBKG)} is constructed by merging the DDI graph and the eKG. Formally, $\mathcal{G}_{BKG} = (\mathcal{V}_{BKG}, \mathcal{E}_{BKG}, \mathcal{R}_{BKG})$, where $\mathcal{V}_{BKG} = \mathcal{V}_{DDI} \cup \mathcal{V}_{KG}$, $\mathcal{R}_{BKG} = \mathcal{R}_{DDI} \cup \mathcal{R}_{KG}$, and $\mathcal{E}_{BKG} = \mathcal{E}_{DDI} \cup \mathcal{E}_{KG}$. This unified graph $\mathcal{G}_{BKG}$ bridges the gap between direct DDI data and its broader biological context, providing a comprehensive framework that can be directly applied to downstream tasks such as DDI prediction.

\subsection{Substructure-Level Molecular Graph Construction}

For each drug $u$, its Simplified Molecular Input Line Entry System (SMILES) representation can be retrieved from publicly available databases such as PubChem \cite{Sunghwan22PubChem}. Using RDKit\footnote{\href{https://github.com/rdkit/rdkit}{https://github.com/rdkit/rdkit}}, the SMILES string is transformed into a \textit{molecular graph} $\mathcal{G}_u^M = (\mathcal{V}_u^M, \mathcal{E}_u^M)$, where $\mathcal{V}_u^M$ represents the set of atoms (nodes) and $\mathcal{E}_u^M$ the set of chemical bonds (edges). This atomic-level molecular graph accurately represents the molecule’s structural composition and atomic connectivity, providing the foundational information for further analysis.

Despite its fine-grained detail, the atomic-level graph lacks interpretability and does not align with how chemists typically reason about molecular structures—through chemically meaningful substructures, such as functional groups and ring systems. To address this limitation, we employ a chemistry-inspired fragmentation strategy that transforms the atomic graph into a substructure-level graph capturing high-order chemical semantics \cite{Zhong202409}. Specifically, we use the Breaking Retrosynthetically Interesting Chemical Substructures (BRICS) algorithm \cite{Degen2008BRICS}, which applies a predefined set of reaction-based rules to identify cleavage sites relevant to retrosynthetic analysis.

\begin{figure}[ht]
  \centering
  \includegraphics[width=\linewidth]{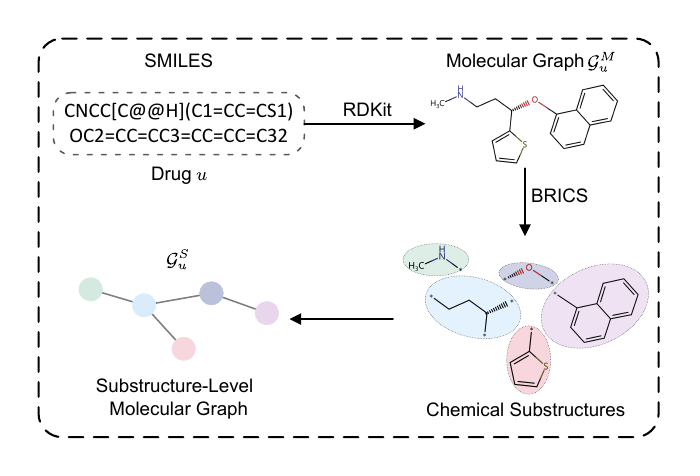}
  \caption [Substructure-level molecular graph of Duloxetine]{The process of constructing a substructure-level molecular graph of a drug molecule (example: Duloxetine\protect\footnotemark).}
  \Description{Substructure-level molecular graph of Duloxetine}
  \label{fig: substructure-level molecular graph.}
\end{figure}

As illustrated in Fig.~\ref{fig: substructure-level molecular graph.}, the BRICS algorithm decomposes the molecular graph $\mathcal{G}_u^M$ into a set of chemically significant substructures. Each substructure—often corresponding to functional groups, ring systems, or other characteristic fragments—is represented as a node in the resulting \textit{substructure-level molecular graph}, $\mathcal{G}_u^S = (\mathcal{V}_u^S, \mathcal{E}_u^S)$. The edges in $\mathcal{E}_u^S$ preserve the connectivity between substructures by incorporating information from the original cleavage points in $\mathcal{G}_u^M$. This representation captures both localized chemical environments and global structural organization, offering a more interpretable and modular view of the molecule.

\footnotetext{A medication used to treat depression, anxiety, nerve pain, and incontinence. \href{https://go.drugbank.com/drugs/DB00476}{DrugBank ID: DB00476.}}

\section{Methodology}

\begin{figure*}[htp]
  \centering
  \includegraphics[width=\textwidth]{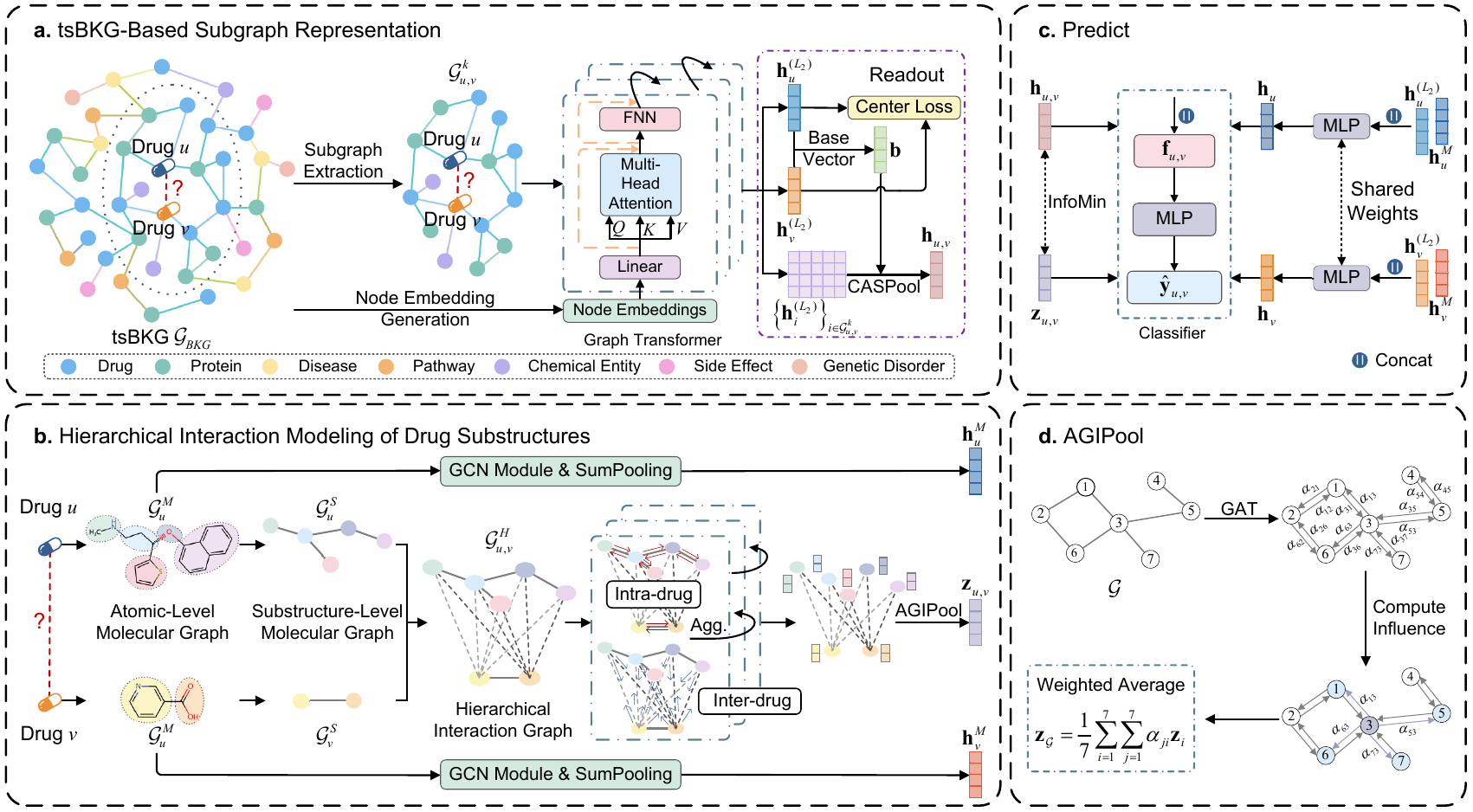}
  \caption{An overview of the proposed MolecBioNet framework (a, b, c) and a schematic illustration of the attention-guided influence pooling (AGIPool) process (d).}
  \Description{Overall Framework}
  \label{fig: Overall Framework}
\end{figure*}

Fig. \ref{fig: Overall Framework} presents the framework of our proposed model, MolecBioNet. The following subsections describe the model's components in detail, outlining the specific methods employed in each module.

\subsection{tsBKG-Based Subgraph Representation}

This subsection focuses on embedding representations of drug pairs within a tsBKG by extracting and encoding local subgraphs. As illustrated in Fig. \ref{fig: Overall Framework}a, the process consists of the following steps: (1) generating initial node embeddings, (2) extracting $k$-hop subgraphs surrounding each drug pair and encoding their features, (3) leveraging a Graph Transformer to perform message passing, and (4) applying a context-aware subgraph pooling mechanism to derive the final representation of the drug pair. Each step is detailed as follows.

\subsubsection{Node Embedding Generation in tsBKG}
To capture global relational patterns and enhance drug node feature representations with contextual information from diverse biomedical entities, a GNN is applied to the tsBKG $\mathcal{G}_{BKG} = (\mathcal{V}_{BKG}, \mathcal{E}_{BKG}, \mathcal{R}_{BKG})$. This process generates embeddings for each entity in the graph. Let $\mathbf{x}_v^{(0)}$ represent the initial feature of node $v \in \mathcal{V}_{BKG}$, which, in this work, is represented as a one-hot vector. Using the GraphSAGE framework \cite{HamiltonGraphSAGE}, the node embedding $\mathbf{x}_v^{(l)}$ is updated iteratively at the $l$-th layer as follows:
\begin{equation}\label{GraphSAGE1}
  \mathbf{x}_{\mathcal{N}(v)}^{(l+1)} = \mathrm{MEAN}\left(\{\mathbf{W}_1^{(l)}\mathbf{x}_u^{(l)} \mid  u \in \mathcal{N}(v)\}\right),
\end{equation}
\begin{equation}\label{GraphSAGE2}
  \mathbf{x}_v^{(l+1)} = \sigma \left(\mathbf{W}_2^{(l)} \left[\mathbf{x}_v^{(l)} \Vert \mathbf{x}_{\mathcal{N}(v)}^{(l+1)}\right]\right),
\end{equation}
where $\mathcal{N}(v)$ denotes the neighbors of $v$, $\mathrm{MEAN}(\cdot)$ aggregates features from neighbors, $\mathbf{W}_1^{(l)}$ and $\mathbf{W}_2^{(l)}$ are learnable weight matrices, $\Vert$ denotes concatenation, and $\sigma$ is the ReLU activation function. After $L_1$ layers, the embedding $\mathbf{x}_v^{(L_1)}$ is the final representation of node $v$.

\subsubsection{Extracting Local Subgraphs for Drug Pairs}
To infer DDIs, we treat each drug pair as a unified entity and analyze its contextual dependencies within the biomedical knowledge graph ($\mathcal{G}_{BKG}$), capturing the macro-scale influences of other entities such as proteins and diseases. These influences propagate through the connections in the graph and diminish with increasing distance from the drug pair \cite{Yu202109}. Hence, we focus on the $k$-hop enclosing subgraph centered on the drug pair to focus on the most contextually relevant region while balancing computational efficiency and the inclusion of meaningful connections. Specifically, Given a drug pair $(u, v)$, the \textit{$k$-hop enclosing subgraph} $\mathcal{G}_{u,v}^k$ is induced from $\mathcal{G}_{BKG}$ by including all nodes $i$ satisfying $\{i \mid d(i, u) \leq k \ or \ d(i, v) \leq k \}$, where $d(i, u)$ represents the shortest path distance between nodes $i$ and $u$. 

Encoding the relative positions of nodes within the subgraph is crucial for understanding how entities influence the central node \cite{ZhangSubgraph}. To achieve this, we assign each node $i$ a position vector, $\mathbf{p}_i=[\text{one-hot}(d'(i,u))\Vert \text{one-hot}(d'(i,v))]$, where the adjusted shortest-path distance $d'(i,u)$ and $d'(i,v)$ are defined follows:
$$d'(i, u) =
\begin{cases}
    d(i, u), & \text{if } d(i, u) < \infty, \\
    \max\{d(j,u)+1 \mid j \in \mathcal{G}_{u,v}^k, d(j,u)<\infty\}, & \text{otherwise}.
\end{cases}
$$
$$d'(i, v) =
\begin{cases}
    d(i, v), & \text{if } d(i, v) < \infty, \\
    \max\{d(j,v)+1 \mid j \in \mathcal{G}_{u,v}^k, d(j,v)<\infty\}, & \text{otherwise}.
\end{cases}
$$
Since positional information is encoded using one-hot vectors, nodes at the same distance share identical position encodings. However, when a node $i$ is disconnected from either $u$ or $v$, directly assigning an infinite distance is not feasible for one-hot encoding. To address this, we assign such nodes a pseudo-distance based on the maximum reachable distance in the subgraph, ensuring that all unreachable nodes fall into a consistent category rather than being arbitrarily assigned different encodings. This approach maintains structural coherence and prevents disconnected nodes from being treated as outliers in the representation space.

Beyond positional information, the functional roles of nodes are equally important. To reflect this, each node $i$ is assigned a categorical feature vector, $\mathbf{c}_i = \text{one-hot}(\text{class}(i))$, where $\text{class}(i)$ represents its entity type, such as protein, disease, or another biomedical category. By augmenting the precomputed embeddings $\mathbf{x}_i^{(L_1)}$ with the positional and categorical encodings, we obtain the feature representation for each node $i$, expressed as $\mathbf{h}_i^{(0)} = [\mathbf{x}_i^{(L_1)} \Vert \mathbf{p}_i \Vert \mathbf{c}_i] \in \mathbb{R}^d$.

\subsubsection{Message Passing via Graph Transformer}
To aggregate features within the extracted $k$-hop subgraph $\mathcal{G}_{u,v}^k$, a Graph Transformer \cite{dwivedi2021graphtransformer} is employed. Unlike traditional GNNs, the Graph Transformer leverages multi-head attention mechanisms to effectively capture both local structural patterns and long-range dependencies, enabling richer modeling of the influence of surrounding biomedical entities on the central drug pair. Node embeddings are updated iteratively as follows:
\begin{equation}\label{GraphTransformer}
  \mathbf{\hat{h}}_i^{(l+1)} = \mathbf{O}^{(l)} \mathop{\Vert}\limits_{n=1}^{N} \sum_{j\in \mathcal{N}(i)}w_{i,j}^{(n,l)}\mathbf{V}^{(n,l)} \mathbf{h}_j^{(l)},
\end{equation}
\begin{equation}\label{ATT}
  w_{i,j}^{(n,l)} = \mathrm{softmax}_j\frac{\mathbf{Q}^{(n,l)} \mathbf{h}_i^{(l)} \cdot \mathbf{K}^{(n,l)} \mathbf{h}_j^{(l)}}{\sqrt{d_n}},
\end{equation}
where $\mathbf{Q}^{(n,l)}$, $\mathbf{K}^{(n,l)}$, and $\mathbf{V}^{(n,l)} \in \mathbb{R}^{d_n \times d}$ are learnable matrices for query, key, and value projections, $d_n$ is the dimensionality of each head, and $\mathrm{softmax}_j$ normalizes attention weights. Multi-head attention provides diverse relational insights, and the outputs of $N$ heads are concatenated and passed through $\mathbf{O}^{(l)} \in \mathbb{R}^{d \times d}$ to maintain dimensional consistency. Then the attention outputs $\mathbf{\hat{h}}_i^{(l+1)}$ are passed through a Feed Forward Network (FFN) for further transformation:
\begin{equation}\label{FNN}
  \mathbf{h}_i^{(l+1)} = \mathbf{W}_4^{(l)}(\mathrm{GELU}(\mathbf{W}_3^{(l)} \mathbf{\hat{h}}_i^{(l+1)})),
\end{equation}
where $\mathbf{W}_3^{(l)}$ and $\mathbf{W}_4^{(l)}$ are learnable weight matrices, $\mathrm{GELU}$ denotes the activation function. After $L_2$ layers, updated embeddings $\mathbf{h}_i^{(L_2)}$ for all nodes are obtained.

\subsubsection{Context-Aware Subgraph Pooling}
To accurately predict DDIs, it is crucial to effectively capture the contextual relationships between a drug pair and its surrounding entities in the biomedical knowledge graph. We propose a \textit{context-aware subgraph pooling (CASPool)} mechanism, which adaptively aggregates node-level information within the subgraph, assigning greater importance to nodes more relevant to the central drug pair $(u, v)$. The result is a unified subgraph-level embedding that represents the influence of the surrounding biomedical context on drug interaction.

The process begins by constructing a base vector, $\mathbf{b} = [\mathbf{h}_u^{(L_2)} \Vert \mathbf{h}_v^{(L_2)}]$, which concatenates the embeddings of the two central drugs derived from the final Graph Transformer layer. Using this base vector, we compute an attention score $\alpha_i$ for each node $i$ in the subgraph. This score quantifies the node's contribution to the representation of the drug pair: 
\begin{equation}\label{attpool}
  \alpha_i = \mathrm{softmax}_i \left(\mathbf{W}_h \mathbf{h}_i^{(L_2)} \cdot \mathbf{W}_b \mathbf{b} \right),
\end{equation}
where $\mathbf{W}_h$ and $\mathbf{W}_b$ are learnable matrices that project node embeddings $\mathbf{h}_i^{(L_2)}$ and the base vector $\mathbf{b}$ into a shared latent space. The softmax function normalizes the scores across all nodes in the subgraph, ensuring $\sum_{i \in \mathcal{G}_{u,v}^k} \alpha_i = 1$.

With these attention scores, we compute the subgraph-level representation by aggregating the re-weighted node embeddings:
\begin{equation}\label{BKG}
    \mathbf{h}_{u,v} = \mathrm{MEAN}\left(\left\{\alpha_i \mathbf{h}_i^{(L_2)}\mid i \in \mathcal{G}_{u,v}^k\right\}\right).
\end{equation}
The resulting pooled embedding, $\mathbf{h}_{u,v}$, serves as the final representation of the drug pair within the knowledge graph, effectively summarizing their contextual interactions and the contributions of surrounding entities.

\subsection{Hierarchical Interaction Modeling of Drug Substructures}
Based on medicinal chemistry knowledge, drugs are complex entities composed of various chemically meaningful substructures, which collectively dictate their pharmacokinetic (how the body processes the drug) and pharmacodynamic (how the drug affects the body) properties \cite{Harrold01072014, TangDSIL-DDI}. Importantly, DDIs often stem from the chemical reactivity of specific substructures rather than the entirety of a drug's molecular structure \cite{huang2020caster, MedFused}. Recognizing this, accurate prediction of DDIs requires an approach that explicitly models the behavior of substructures, capturing intra-drug relationships alongside inter-drug interactions.

To address this need, we present a \textit{substructure-aware hierarchical interaction model} that incorporates internal drug structures with cross-drug substructural dependencies. This model bridges the gap between chemical intuition and computational graph modeling, offering a fine-grained and biologically interpretable framework for DDI prediction. The details are described below.

\subsubsection{Modeling Inter- and Intra-Drug Interactions through Message Passing}
For a drug pair $(u, v)$, we construct substructure-level molecular graphs $\mathcal{G}_u^S$ and $\mathcal{G}_v^S$ for the individual drugs. Each node represents a chemical substructure obtained through BRICS decomposition, and its initial features $\mathbf{z}_i^0$ are encoded as Morgan Fingerprints \cite{Morgan} generated from the substructure’s SMILES representation using RDKit. To capture inter-drug substructural interactions, we extend these molecular graphs to form a \textit{hierarchical interaction graph (HIG)} $\mathcal{G}_{u,v}^H$, which connects all nodes in $\mathcal{G}_u^S$ with all nodes in $\mathcal{G}_v^S$. This graph offers the dual perspectives of localized structural dependencies and global inter-drug relationships, enabling the model to encode the intricate patterns underlying DDIs.

The graph's intra-drug edges are processed through a Graph Convolutional Network (GCN) \cite{kipf2017semisupervised} to propagate information between neighboring substructures. This step captures their local structural and chemical dependencies. The node embeddings are updated iteratively as follows:
\begin{equation}\label{gcn}
    \mathbf{z}_{i,\mathrm{intra}}^{(l+1)} = \sigma\left(\sum_{j \in \mathcal{N}(i)} \frac{1}{\sqrt{d_i}\sqrt{d_j}}\mathbf{W}_{\mathrm{intra}}^{(l)} \mathbf{z}_{j,\mathrm{intra}}^{(l)}\right),
\end{equation}
where $\mathcal{N}(i)$ denotes the neighbors of node $i$, $d_i$ represents the degree of node $i$, $\mathbf{W}_{\mathrm{intra}}^{(l)}$ is a learnable weight matrix, and $\sigma$ denotes the ReLU activation function. This yields intra-drug embeddings $\mathbf{z}_{i,\mathrm{intra}}^{(l+1)}$, encoding localized structural and functional characteristics of each substructure.

For inter-drug interactions, we employ a multi-head Graph Attention Network (GAT) \cite{veličković2018graph} on the cross-drug edges of $\mathcal{G}_{u,v}^H$, to adaptively allocate weights to neighboring nodes. The updated node embeddings are computed as:
\begin{equation}\label{gat}
    \mathbf{z}_{i,\mathrm{inter}}^{(l+1)} = \sigma\left(\sum_{j \in \mathcal{M}(i)} \alpha_{ij}^{(l)} \mathbf{W}_{\mathrm{inter}}^{(l)} \mathbf{z}_{j,\mathrm{inter}}^{(l)}\right),
\end{equation}
where $\mathcal{M}(i)$ represents the set of cross-drug neighbors of node $i$, $\mathbf{W}_{\mathrm{inter}}^{(l)}$ is a learnable weight matrix, and $\alpha_{ij}^{(l)}$ is the attention coefficient from node $j$ to $i$, defined as:
\begin{equation}\label{weight}
    \alpha_{ij}^{(l)} = \frac{\mathrm{exp}\left( \mathrm{LeakyReLU} \left(\mathbf{a}^T \left[ \mathbf{W}_{\mathrm{inter}}^{(l)} \mathbf{z}_{i,\mathrm{inter}}^{(l)} \Vert \mathbf{W}_{\mathrm{inter}}^{(l)} \mathbf{z}_{j,\mathrm{inter}}^{(l)} \right] \right)\right)}{\sum_{k \in \mathcal{M}(i)}\mathrm{exp}\left( \mathrm{LeakyReLU} \left(\mathbf{a}^T \left[ \mathbf{W}_{\mathrm{inter}}^{(l)} \mathbf{z}_{i,\mathrm{inter}}^{(l)} \Vert \mathbf{W}_{\mathrm{inter}}^{(l)} \mathbf{z}_{k,\mathrm{inter}}^{(l)} \right] \right)\right)},
\end{equation}
with $\mathbf{a}$ is the learnable parameter vector. 

Finally, the embeddings from intra-drug and inter-drug interactions are combined at each layer:
\begin{equation}
    \mathbf{z}_{i}^{(l+1)} = \mathbf{z}_{i,\mathrm{intra}}^{(l+1)} + \mathbf{z}_{i,\mathrm{inter}}^{(l+1)},
\end{equation}
integrating both localized structural features and cross-drug interaction signals.

\subsubsection{Attention-Guided Influence Pooling}
To obtain a drug-pair-level representation from the hierarchical graph $\mathcal{G}_{u,v}^H$, we introduce \textit{attention-guided influence pooling (AGIPool)}. As shown in Fig. \ref{fig: Overall Framework}d, AGIPool leverages the attention coefficients $\alpha_{ij}$ computed by the Graph Attention Network to quantify the relative influence of each node. Intuitively, a node that receives more attention from its neighbors is considered more influential, as it contributes more significantly to their representation updates \cite{Gu2021rank}. This aligns with the underlying chemistry of drug interactions, where certain functional substructures disproportionately contribute to molecular behavior and interactions. Mathematically, the influence score $c_i$ for node $i$ is defined as:

\begin{equation}
    c_i = \sum_{j\in \mathcal{N}(i)}\alpha_{ji}^{(L_3)},
\end{equation}
where $\alpha_{ji}^{(L_3)}$ represents the attention weight assigned by node $j$ to node $i$ in the final layer $L_3$ of message passing process. For simplicity, we use a single attention head in the final GAT layer.

The final drug-pair embedding $\mathbf{z}_{u,v}$ is obtained by aggregating the node embeddings, weighted by their influence scores:
\begin{equation}
    \mathbf{z}_{u,v} = \mathrm{MEAN} \left( \left\{c_i\mathbf{z}_i^{(L_3)} | i \in \mathcal{G}_{u,v}^H \right\} \right).
\end{equation}
This pooling approach ensures that the resulting representation emphasizes the most influential substructures, enhancing interpretability and predictive accuracy. By integrating domain knowledge of chemically meaningful substructures with classical graph neural networks, this substructure-aware hierarchical interaction model offers a robust and interpretable framework for DDI prediction at the molecular level.

\begin{table*}
    \begin{threeparttable}
        \caption{Performance comparison on Ryu's dataset and DrugBank dataset. All results are obtained through 5-fold cross-validation, with the mean and standard deviation reported on the test set. Bold values indicate the best performance. P-values are computed using the corrected paired Student's t-test \cite{paired_t-test} comparing MolecBioNet with the second-best model, TIGER.}
        \label{tab:dataset-comparison}
        \centering
        \begin{tabularx}{\textwidth}{l *{8}{>{\centering\arraybackslash}X}}
        \toprule
        \multirow{2}{*}{\textbf{Methods}} & \multicolumn{4}{c}{\textbf{Ryu's Dataset}} & \multicolumn{4}{c}{\textbf{DrugBank Dataset}} \\
        \cmidrule(lr){2-5} \cmidrule(lr){6-9}
        & \textbf{ACC} & \textbf{F1} & \textbf{PR-AUC} & \textbf{Cohen's $\kappa$} & \textbf{ACC} & \textbf{F1} & \textbf{PR-AUC} & \textbf{Cohen's $\kappa$} \\
        \midrule
        DeepDDI \cite{Ryu2018DeepDDI} & 0.868$\pm$0.005 & 0.703$\pm$0.006 & 0.811$\pm$0.010 & 0.842$\pm$0.006 & 0.862$\pm$0.003 & 0.676$\pm$0.026 & 0.810$\pm$0.016 & 0.848$\pm$0.004 \\
        SSI-DDI \cite{Nyamabo202111} & 0.848$\pm$0.006 & 0.618$\pm$0.017 & 0.704$\pm$0.014 & 0.818$\pm$0.007 & 0.743$\pm$0.009 & 0.327$\pm$0.005 & 0.438$\pm$0.009 & 0.717$\pm$0.010 \\
        GGI-DDI \cite{Yu202404} & 0.889$\pm$0.006 & 0.811$\pm$0.019 & 0.880$\pm$0.012 & 0.869$\pm$0.007 & 0.859$\pm$0.008 & 0.762$\pm$0.015 & 0.836$\pm$0.016 & 0.846$\pm$0.009 \\
        KGNN \cite{KGNNijcai2020} & 0.916$\pm$0.003 & 0.814$\pm$0.021 & 0.872$\pm$0.014 & 0.900$\pm$0.004 & 0.907$\pm$0.002 & 0.764$\pm$0.011 & 0.832$\pm$0.007 & 0.898$\pm$0.002 \\
        SumGNN \cite{Yu202109} & 0.904$\pm$0.002 & 0.842$\pm$0.005 & 0.912$\pm$0.013 & 0.886$\pm$0.002 & 0.864$\pm$0.002 & 0.732$\pm$0.019 & 0.855$\pm$0.011 & 0.851$\pm$0.002 \\
        LaGAT \cite{LaGATHong202212} & 0.935$\pm$0.001 & 0.843$\pm$0.014 & 0.899$\pm$0.012 & 0.922$\pm$0.001 & 0.900$\pm$0.001 & 0.741$\pm$0.010 & 0.829$\pm$0.008 & 0.891$\pm$0.001 \\
        KnowDDI \cite{Wang202403} & 0.923$\pm$0.002 & 0.880$\pm$0.010 & 0.934$\pm$0.011 & 0.909$\pm$0.002 & 0.894$\pm$0.001 & 0.772$\pm$0.015 & 0.868$\pm$0.008 & 0.885$\pm$0.001 \\
        TIGER \cite{Su202403} & 0.938$\pm$0.003 & 0.880$\pm$0.007 & 0.894$\pm$0.011 & 0.926$\pm$0.004 & 0.917$\pm$0.001 & 0.791$\pm$0.008 & 0.763$\pm$0.011 & 0.910$\pm$0.001 \\
        MolecBioNet & \textbf{0.952$\pm$0.002} & \textbf{0.912$\pm$0.008} & \textbf{0.942$\pm$0.006} & \textbf{0.943$\pm$0.003}& \textbf{0.935$\pm$0.002} & \textbf{0.846$\pm$0.010} & \textbf{0.910$\pm$0.008} & \textbf{0.929$\pm$0.002} \\
        \midrule
        P-values & $7.1 \times 10^{-3}$ & $1.4 \times 10^{-2}$ & $7.2 \times 10^{-3}$ & $7.1 \times 10^{-3}$ & $3.6 \times 10^{-4}$ & $2.0 \times 10^{-3}$ & $1.1 \times 10^{-5}$ & $3.4 \times 10^{-4}$ \\
        \bottomrule
        \end{tabularx}
    \end{threeparttable}
\end{table*}

\subsection{Multi-Scale Embedding Fusion and Loss Design}

To ensure the embeddings derived from the biological network-level $k$-hop subgraph $\mathcal{G}_{u,v}^k$ (\(\mathbf{h}_{u,v}\)) and the molecular-level hierarchical interaction graph $\mathcal{G}_{u,v}^H$ (\(\mathbf{z}_{u,v}\)) are complementary while minimizing redundancy, we introduce a \textit{mutual information (MI) minimization} \cite{Zhang_2021_ICCV} regularization term. Mutual information measures the dependence between two variables by quantifying how much uncertainty in one variable is reduced given knowledge of the other. For the pairwise embeddings \(\mathbf{h}_{u,v}\) and \(\mathbf{z}_{u,v}\), MI is defined as:
\begin{equation}
    MI(\mathbf{h}_{u,v}, \mathbf{z}_{u,v}) = H(\mathbf{h}_{u,v}) + H(\mathbf{z}_{u,v}) - H(\mathbf{h}_{u,v}, \mathbf{z}_{u,v}),
\end{equation}
where \(H(\cdot)\) represents entropy and \(H(\mathbf{h}_{u,v}, \mathbf{z}_{u,v})\) is the joint entropy. Intuitively, a high $MI$ value indicates significant shared information between the two embeddings, which is undesirable in this context as it indicates redundancy.

Since directly computing \(H(\mathbf{h}_{u,v})\) and \(H(\mathbf{z}_{u,v})\) is challenging, we reformulate \(MI(\mathbf{h}_{u,v}, \mathbf{z}_{u,v})\) using conditional entropy and Kullback-Leibler (KL) divergence:
\begin{equation}
\begin{aligned}
    MI(\mathbf{h}_{u,v}, \mathbf{z}_{u,v}) = & H_{\mathbf{z}_{u,v}}(\mathbf{h}_{u,v}) + H_{\mathbf{h}_{u,v}}(\mathbf{z}_{u,v}) -H(\mathbf{h}_{u,v}, \mathbf{z}_{u,v})\\
    & - KL(\mathbf{h}_{u,v} \Vert \mathbf{z}_{u,v}) - KL(\mathbf{z}_{u,v} \Vert \mathbf{h}_{u,v}),
\end{aligned}
\end{equation}
\begin{equation}
    KL(\mathbf{h}_{u,v} \Vert \mathbf{z}_{u,v}) = H_{\mathbf{z}_{u,v}}(\mathbf{h}_{u,v}) - H(\mathbf{h}_{u,v}),
\end{equation}
\begin{equation}
    KL(\mathbf{z}_{u,v} \Vert \mathbf{h}_{u,v}) = H_{\mathbf{h}_{u,v}}(\mathbf{z}_{u,v}) - H(\mathbf{z}_{u,v}),
\end{equation}
where \(H_{\mathbf{z}_{u,v}}(\mathbf{h}_{u,v}) = -\mathbb{E}[\log p(\mathbf{h}_{u,v} \mid \mathbf{z}_{u,v})]\) is the cross-entropy of \(\mathbf{h}_{u,v}\) conditioned on \(\mathbf{z}_{u,v}\). This reformulation allows efficient computation and optimization during training. Minimizing this MI term constrains the embeddings \(\mathbf{h}_{u,v}\) and \(\mathbf{z}_{u,v}\) to capture unique and complementary aspects of the biological and molecular interactions.

Since $H(\mathbf{h}_{u,v}, \mathbf{z}_{u,v})$ is non-negative, the mutual information loss is defined as:
\begin{equation}
\begin{aligned}
\mathcal{L}_{MI} = \mathbb{E}_{(u, v) \sim \mathcal{D}} \Big[ 
    & H_{\mathbf{z}_{u,v}}(\mathbf{h}_{u,v}) + H_{\mathbf{h}_{u,v}}(\mathbf{z}_{u,v}) \\
    & \quad - KL(\mathbf{h}_{u,v} \Vert \mathbf{z}_{u,v}) - KL(\mathbf{z}_{u,v} \Vert \mathbf{h}_{u,v})
\Big],
\end{aligned}
\end{equation}
where \(\mathcal{D}\) denotes the dataset or a mini-batch of samples.

For each drug \(u\), we compute a biological embedding \(\mathbf{h}_u^{(L_2)}\) derived from the \(k\)-hop subgraph $\mathcal{G}_u^k$ and a molecular embedding \(\mathbf{h}_u^M\) obtained from its molecular graph \(\mathcal{G}_u^M\) via a $L_4$-layers GCN (more details in Appendix \ref{Molecular Graph Representation}). These embeddings are fused through a multi-layer perceptron (MLP):
\begin{equation}
\mathbf{h}_u = \mathrm{MLP}\left(\left[\mathbf{h}_u^{(L_2)} \Vert \mathbf{h}_u^M \right]\right).
\end{equation}
This combined embedding $\mathbf{h}_u$ captures both the biological and molecular features of the drug $u$, making it suitable for further drug interaction analysis. 

For each drug pair \((u,v)\), we construct a joint representation that integrates biological, molecular, and pairwise features:
\begin{equation}
\mathbf{f}_{u,v} = \left[\mathbf{h}_{u,v} \Vert \mathbf{z}_{u,v} \Vert \mathbf{h}_u \Vert \mathbf{h}_v\right].
\end{equation}
A classifier is then used to predict the interaction type between the pair:
\begin{equation}
    \mathbf{\hat{y}}_{u,v} = \mathbf{W}_{c} \mathbf{f}_{u,v},
\end{equation}
where $\mathbf{W}_{c}$ is the learnable weight matrix of the classifier.

We frame DDI prediction as a multi-class classification task and optimize using a cross-entropy loss:
\begin{equation}
    \mathcal{L}_{P} = - \sum_{(u,r,v) \in \mathcal{E}_{DDI}} \mathbf{y}_{u,v}^T \log \mathbf{\hat{y}}_{u,v},
\end{equation}
where $\mathbf{y}_{u,v}$ is the one-hot encoded ground truth vector corresponding to the drug interaction type.

To account for the shared nature of drugs across multiple pairs, we employ a \textit{center loss} \cite{WenCenterLoss} that constrains the embeddings of the same drug to be similar across different pairs:
\begin{equation}
    \mathcal{L}_{C} = \frac{1}{2} \sum_{(u,r,v) \in \mathcal{E}_{DDI}} \Big( \Vert \mathbf{h}_u^{(L_2)} - \mathbf{c}_u \Vert_2^2 + \Vert \mathbf{h}_v^{(L_2)} - \mathbf{c}_v \Vert_2^2 \Big),
\end{equation}
where \(\mathbf{c}_u\) and \(\mathbf{c}_v\) are learnable center vectors for drugs \(u\) and \(v\), respectively.

The overall loss function combines the prediction loss, center loss, and mutual information minimization loss:
\begin{equation}
    \mathcal{L} = \mathcal{L}_{P} + \beta \mathcal{L}_{C} + \gamma \mathcal{L}_{MI},
\end{equation}
where \(\beta\) and \(\gamma\) are hyperparameters controlling the contributions of center loss and mutual information minimization loss.

\section{Results}

We used two benchmark datasets for multi-class DDI prediction: Ryu's dataset \cite{Ryu2018DeepDDI} and an updated version of the DrugBank dataset (version 5.1.12, March 14, 2024) \cite{DrugBank}. To ensure robust evaluation, we applied 5-fold cross-validation and measured performance using four metrics: accuracy (ACC), macro-averaged F1 Score (F1), PR-AUC, and Cohen's Kappa. All experimental settings, including detailed descriptions of the datasets, baseline configurations, novel drug experimental setups, ablation study design, and evaluation metrics, as well as the associated datasets and source code for MolecBioNet, are publicly available at: \href{https://github.com/MengjieChan/MolecBioNet}{https://github.com/Mengjie
Chan/MolecBioNet}.

\subsection{Analysis of Model Performance}

We evaluate MolecBioNet against several state-of-the-art models on Ryu’s and DrugBank datasets, with results summarized in Table \ref{tab:dataset-comparison}. Across all evaluation metrics, MolecBioNet consistently outperforms competing approaches, highlighting the effectiveness of hierarchical molecular interaction modeling and context-aware representation learning from biomedical knowledge graphs. These findings reinforce the necessity of integrating both molecular-level interactions and knowledge-driven contextual relationships for accurate multi-class DDI prediction.

Among the competing methods, TIGER \cite{Su202403} leverages both biomedical knowledge graphs and molecular graphs, leading to better performance than models relying solely on biomedical knowledge (KGNN \cite{KGNNijcai2020}, SumGNN \cite{Yu202109}, LaGAT \cite{LaGATHong202212}, KnowDDI \cite{Wang202403}) or molecular structures (SSI-DDI \cite{Nyamabo202111}, GGI-DDI \cite{Yu202404}). However, TIGER’s approach to representation learning is inherently limited. Rather than explicitly capturing interaction mechanisms, it encodes each drug separately and then concatenates their embeddings to form a drug-pair representation. This naïve fusion strategy overlooks crucial biochemical and pharmacological interactions, ultimately restricting its predictive power.

For molecular graph-based models (SSI-DDI, GGI-DDI), they treat drug-drug interaction types as contextual inputs rather than direct prediction targets, effectively reducing the problem to a binary classification task (i.e., determining whether a given drug pair exhibits a specific pre-defined interaction). When reformulated as a true multi-class classification problem, their performance is likely to degrade due to the lack of explicit interaction modeling. This limitation is particularly evident in SSI-DDI, which performs even worse than DeepDDI \cite{Ryu2018DeepDDI}—a model that relies solely on molecular fingerprints to infer interactions based on structural similarity.

\subsection{Effectiveness on Novel Drugs}

\begin{table*}
    \begin{threeparttable}
        \caption{Performance comparison of MolecBioNet with baseline models in the Novel Drug–Existing Drug and Novel Drug–Novel Drug settings. All results are obtained through 5-fold cross-validation, with the mean and standard deviation reported on the test set. Bold values indicate the best performance.}
        \label{tab: novel drug results}
        \centering
        \begin{tabularx}{\textwidth}{l *{8}{>{\centering\arraybackslash}X}}
        \toprule
        \multirow{2}{*}{\textbf{Methods}} & \multicolumn{4}{c}{\textbf{Novel Drug–Existing Drug}} & \multicolumn{4}{c}{\textbf{Novel Drug–Novel Drug}} \\
        \cmidrule(lr){2-5} \cmidrule(lr){6-9}
        & \textbf{ACC} & \textbf{F1} & \textbf{PR-AUC} & \textbf{Cohen's $\kappa$} & \textbf{ACC} & \textbf{F1} & \textbf{PR-AUC} & \textbf{Cohen's $\kappa$} \\
        \midrule
        DeepDDI \cite{Ryu2018DeepDDI} & 0.630$\pm$0.028 & 0.459$\pm$0.018 & 0.430$\pm$0.021 & 0.552$\pm$0.031 & 0.372$\pm$0.028 & 0.197$\pm$0.018 & 0.144$\pm$0.017 & 0.238$\pm$0.021 \\
        SSI-DDI \cite{Nyamabo202111} & 0.569$\pm$0.020 & 0.360$\pm$0.021 & 0.357$\pm$0.030 & 0.477$\pm$0.020 & 0.337$\pm$0.048 & 0.144$\pm$0.017 & 0.125$\pm$0.015 & 0.181$\pm$0.032 \\
        GGI-DDI \cite{Yu202404} & 0.636$\pm$0.023 & 0.486$\pm$0.024 & 0.485$\pm$0.030 & 0.559$\pm$0.022 & 0.364$\pm$0.037 & 0.170$\pm$0.008 & 0.153$\pm$0.012 & 0.239$\pm$0.024 \\
        KGNN \cite{KGNNijcai2020} & 0.418$\pm$0.029 & 0.044$\pm$0.008 & 0.069$\pm$0.014 & 0.268$\pm$0.023 & 0.207$\pm$0.034 & 0.018$\pm$0.001 & 0.172$\pm$0.026 & 0.047$\pm$0.020 \\
        SumGNN \cite{Yu202109} & 0.580$\pm$0.027 & 0.405$\pm$0.025 & 0.421$\pm$0.033 & 0.486$\pm$0.027 & 0.311$\pm$0.058 & 0.078$\pm$0.017 & 0.093$\pm$0.024 & 0.102$\pm$0.040 \\
        LaGAT \cite{LaGATHong202212} & 0.557$\pm$0.015 & 0.249$\pm$0.011 & 0.276$\pm$0.012 & 0.455$\pm$0.012 & 0.227$\pm$0.044 & 0.012$\pm$0.001 & 0.171$\pm$0.025 & 0.002$\pm$0.021 \\
        KnowDDI \cite{Wang202403} & 0.516$\pm$0.023 & 0.282$\pm$0.014 & 0.269$\pm$0.014 & 0.409$\pm$0.017 & 0.264$\pm$0.064 & 0.015$\pm$0.001 & 0.017$\pm$0.004 & 0.046$\pm$0.014 \\
        TIGER \cite{Su202403} & 0.543$\pm$0.019 & 0.322$\pm$0.028 & 0.266$\pm$0.026 & 0.442$\pm$0.018 & 0.269$\pm$0.020 & 0.037$\pm$0.010 & 0.037$\pm$0.007 & 0.071$\pm$0.019 \\
        MolecBioNet & \textbf{0.652$\pm$0.020} & \textbf{0.535$\pm$0.017} & \textbf{0.549$\pm$0.039} & \textbf{0.576$\pm$0.021}& \textbf{0.388$\pm$0.049} & \textbf{0.208$\pm$0.019} & \textbf{0.179$\pm$0.029} & \textbf{0.240$\pm$0.041} \\
        \bottomrule
        \end{tabularx}
    \end{threeparttable}
\end{table*}

\begin{figure*}[h]
  \centering

  \subfigure{
    \includegraphics[width=0.48\textwidth]{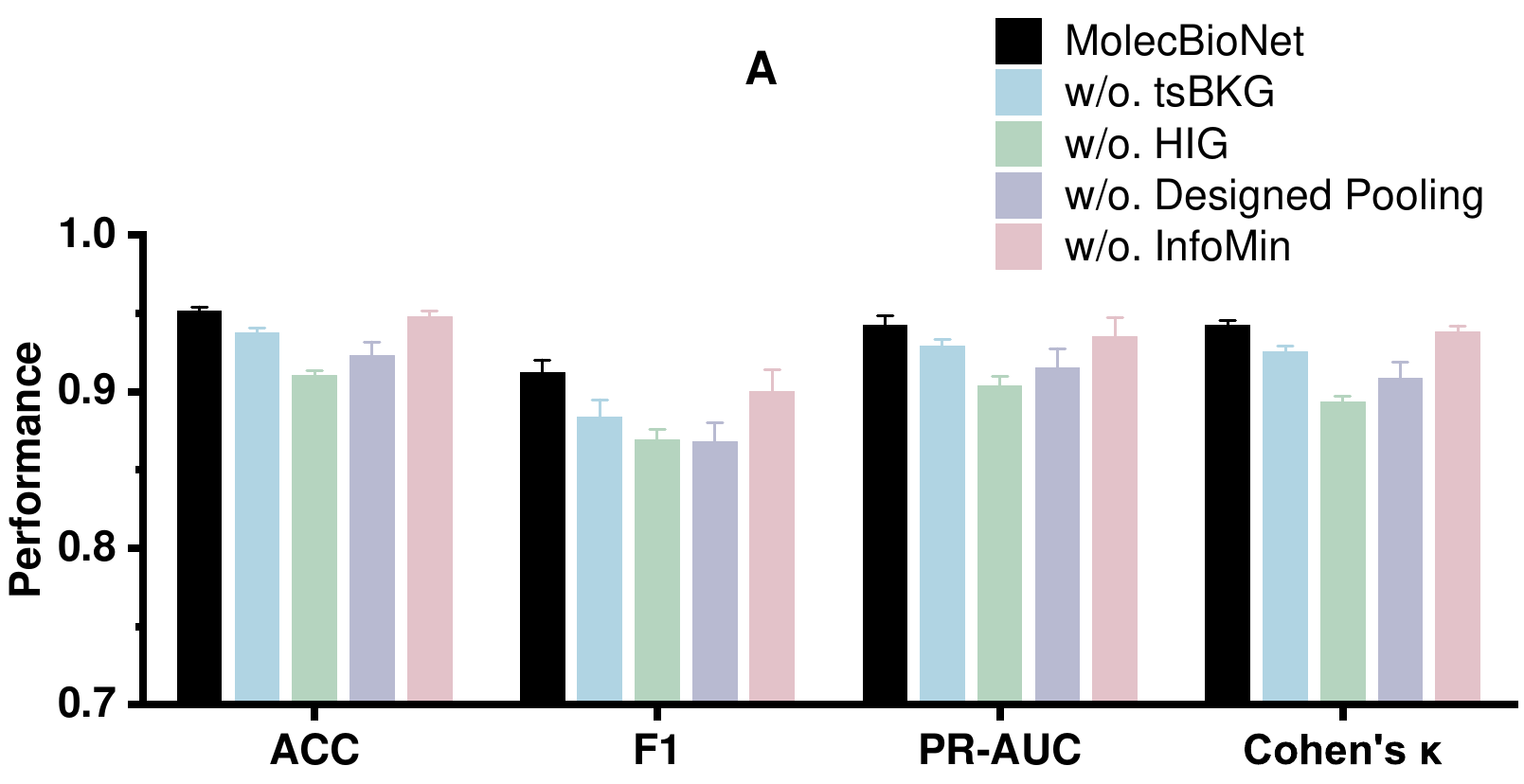}
    \label{fig:subfig1}
  }
  \hfill
  \subfigure{
    \includegraphics[width=0.48\textwidth]{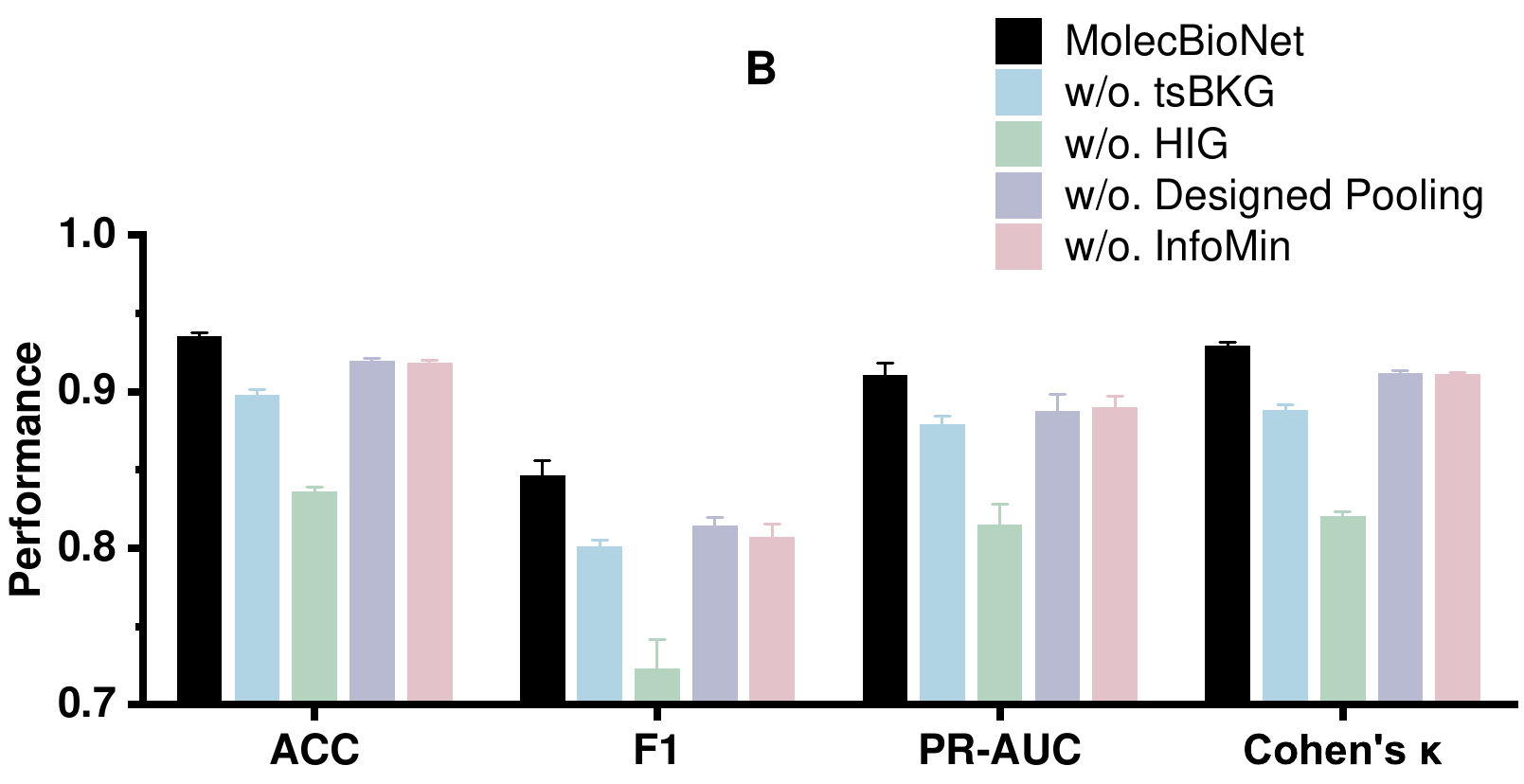}
    \label{fig:subfig2}
  }

  \caption{Performance of MolecBioNet and its variants on the two datasets. (A) Performance for the Ryu's dataset. (B) Performance for the DrugBank dataset.}
  \label{fig: Ablation Study.}
\end{figure*}

To evaluate MolecBioNet’s ability to predict drug interactions involving novel drugs, we conduct two experiments using Ryu’s dataset: Novel Drug–Existing Drug and Novel Drug–Novel Drug. These experiments aim to predict interactions between new drugs and existing drugs, as well as between novel drugs themselves. As shown in Table \ref{tab: novel drug results}, MolecBioNet consistently outperforms the baseline models in both experimental settings. This demonstrates that by incorporating both biomedical knowledge graph relationships and molecular graph information, MolecBioNet is highly effective at learning the relationships between novel drugs and other drugs, even in a cold-start scenario.

The baseline models, including DeepDDI, which uses molecular fingerprints, and SSI-DDI and GGI-DDI, which rely on molecular graphs, perform well in cold-start conditions. However, their performance is still limited, as they lack the biological context provided by biomedical knowledge graphs. In contrast, models that rely solely on biomedical knowledge graphs, such as KGNN, SumGNN, LaGAT and KnowDDI, show significantly poorer performance, particularly in the Novel Drug–Novel Drug experiment. This is primarily because these models are unable to incorporate molecular-level information, leading to inaccurate predictions, especially when novel drugs have limited relationships with other entities in the biomedical knowledge graph. Additionally, while TIGER incorporates both biomedical knowledge graphs and molecular graphs, its performance in novel drug experiments does not meet expectations, further validating the importance of modeling drug pairs as unified entities in our approach.

\begin{table*}[ht]
\centering
\begin{tabularx}{\textwidth}{>{\centering\arraybackslash}X >{\centering\arraybackslash}X >{\centering\arraybackslash}X >{\centering\arraybackslash}X >{\centering\arraybackslash}X  >{\centering\arraybackslash}X }
  \toprule
  Sparsity & 0.5 & 0.6 & 0.7 & 0.8 & 0.9 \\
  \midrule
  $Fidelity_+$ & 0.282±0.014 & 0.252±0.016 & 0.231±0.017 & 0.217±0.015 & 0.213±0.016 \\
  $Fidelity_-$ & 0.269±0.024 & 0.297±0.023 & 0.316±0.022 & 0.325±0.020 & 0.330±0.020 \\
  \bottomrule
\end{tabularx}
\caption{Interpretability evaluation via Fidelity scores under varying sparsity levels.}
\label{tab: metrics of interpretability.}
\end{table*}

\subsection{Ablation Study}

To assess the contribution of each component in MolecBioNet, we conduct ablation studies by removing specific modules and evaluating their impact on model performance. The results, presented in Fig. \ref{fig: Ablation Study.}, highlight the critical role of each component in improving DDI prediction accuracy.

Removing biomedical knowledge-based subgraph representations (w/o. tsBKG) or hierarchical interaction graphs (w/o. HIG) significantly reduces performance, demonstrating the necessity of capturing both macro-level biological interactions and micro-level molecular influences. Notably, the removal of molecular-level interaction modeling (w/o. HIG) causes a more pronounced decline, emphasizing the importance of substructure-aware interaction learning in accurately modeling DDIs. Additionally, replacing CASPool and AGIPool with MeanPooling (w/o. Designed Pooling) leads to performance degradation, confirming that context-aware relational analysis in biomedical subgraphs and substructure-guided interaction prioritization are essential for learning rich and discriminative drug pair representations. Furthermore, removing the mutual information minimization loss term (w/o InfoMin) results in a consistent decline in performance, indicating its effectiveness in enhancing representation quality by reducing redundant information. It also contributes to improved generalization, as evidenced by the lower generalization error and higher generalization ratio of models trained with InfoMin, as shown in Table~\ref{tab: generalization ability of infomin}, especially on the larger DrugBank dataset. These findings collectively affirm the significance of InfoMin in learning robust and generalizable representations.

\subsection{Interpretability and Explanation of Results}

Interpretability is essential for understanding the mechanisms underlying DDIs, in addition to achieving accurate predictions. To quantitatively evaluate the interpretability of MolecBioNet, we adopt the Fidelity \cite{amara2022graphframex} and Sparsity \cite{TPAMI_Explainability} metrics, which assess the faithfulness and compactness of model explanations. As shown in Table~\ref{tab: metrics of interpretability.}, increasing sparsity—i.e., reducing the number of explanatory nodes—leads to a gradual decrease in $Fidelity_+$, suggesting that the explanations gradually become less necessary, which aligns with expectations. However, the smooth trend indicates that removing more explanatory nodes does not significantly degrade explanation quality. Meanwhile, $Fidelity_-$ remains consistently low across all sparsity levels, showing that even a small subset of critical nodes (e.g., the top 10\%) is sufficient to maintain predictive performance. Notably, during the evaluation, only node features are masked, while the graph structure remains intact. The relatively low $Fidelity_+$ values emphasize the importance of structural information in model reasoning, further validating the effectiveness of our graph-based design.

\begin{table}[ht]
\centering
\resizebox{\linewidth}{!}{
\begin{tabular}{>{\centering\arraybackslash}m{1.1cm}>{\centering\arraybackslash}m{2.1cm}c>{\centering\arraybackslash}m{1.2cm}>{\centering\arraybackslash}m{1.2cm}c}
\hline
\multicolumn{2}{c}{\textbf{Chemical Substructure}} & \textbf{\# Drug Pairs} & \textbf{\# Hits (Top 1)} & \textbf{\# Hits (Top 2)} & \textbf{\# Drugs} \\
\hline
\includegraphics[height=0.8cm]{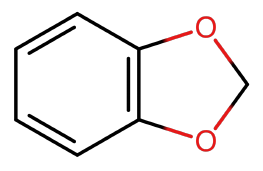} & 1,3-Benzdioxole & 1,417 & 120 & 767 & 3 \\
\includegraphics[height=0.22cm]{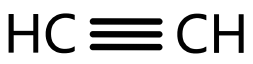} & Acetylene & 425 & 337 & 40 & 3 \\
\includegraphics[height=0.65cm]{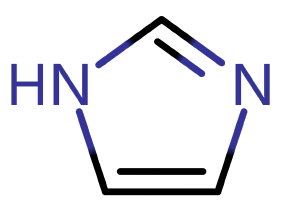} & Imidazole & 1,110 & 94 & 599 & 2 \\
\includegraphics[height=0.26cm]{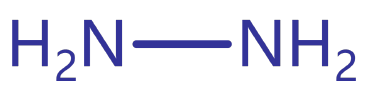} & Hydrazine & 839 & 546 & 293 & 2 \\
\includegraphics[height=0.8cm]{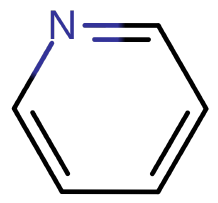} & Pyridine & 705 & 184 & 84 & 2 \\
\includegraphics[height=0.8cm]{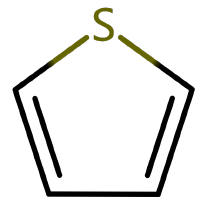} & Thiophene & 1,348 & 29 & 379 & 3 \\
\includegraphics[height=0.8cm]{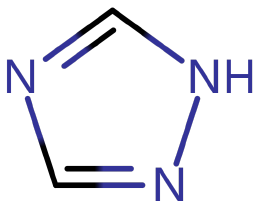} & Triazole & 2,495 & 821 & 1,369 & 4 \\
\hline
\end{tabular}
}
\caption{Literature-supported chemical substructures identified as most influential in DDIs by MolecBioNet.}
\label{tab: chemical substructure.}
\end{table}

To further examine the model’s ability to provide biologically meaningful explanations, we apply the model to a curated dataset of 8,339 drug pairs involving 19 drugs, each known to participate in metabolism-related DDIs via specific chemical substructures \cite{Zhong202409, Hakkola2020}. Using the model’s attention-guided influence pooling, we identify the two most influential chemical substructures for each interaction. As summarized in Table~\ref{tab: chemical substructure.}, the top-1 predicted substructure matches literature-reported functional groups in 2,131 DDIs, while at least one of the top-2 predictions aligns in 5,662 DDIs, highlighting strong alignment between model explanations and domain knowledge.

We also conduct a case study on the interaction between Itraconazole and Promazine, where Itraconazole is known to inhibit the metabolism of Promazine \cite{Lynch2007, DrugBank}. MolecBioNet identifies three key substructures in Itraconazole contributing to this interaction (Fig.~\ref{fig: Itraconazole.}), with Triazole ranked highest. This aligns with evidence that Triazole binds to CYP3A4 and inhibits its enzymatic activity, thereby affecting Promazine metabolism \cite{Hakkola2020}. Moreover, by extracting the local subgraph of Itraconazole and Promazine from tsBKG and applying context-aware subgraph pooling, we highlight the top 10\% most influential nodes, visualized in Fig.~\ref{fig: explain.}. The visualization shows that Itraconazole inhibits the ABCG2 transporter, which may affect the metabolism of other drugs such as Atogepant and Mycophenolate mofetil through modulation of ABCG2-mediated transport \cite{DrugBank}, offering additional insights into its interaction mechanism.

These results collectively demonstrate that MolecBioNet not only achieves high predictive performance but also delivers faithful and biologically grounded explanations across multiple interpretability levels.

\section{Conclusion}

In this work, we introduce MolecBioNet, a graph-based framework for predicting DDIs by integrating molecular interactions with biomedical knowledge graphs. Our results show that MolecBioNet outperforms existing models, capturing complex drug pair relationships through hierarchical modeling and context-aware learning. The model also performs well in cold-start scenarios, effectively predicting interactions for novel drugs. The interpretability of MolecBioNet is further demonstrated through its pooling mechanisms, which provide insights into the key chemical substructures and contextual relationships driving DDIs.

\begin{acks}
This work was supported in part by the National Natural Science Foundation of China under Grant 12471488 and Grant 12231018 and in part by the Postdoctoral Fellowship Program of CPSF under Grant GZC20232912.
\end{acks}

\bibliographystyle{ACM-Reference-Format}
\bibliography{Reference}

\appendix

\section{Molecular Graph Representation via GCN}\label{Molecular Graph Representation}
To compute the graph-level representation of each drug molecule, we apply a GCN \cite{kipf2017semisupervised} on its molecular graph $\mathcal{G}_u^M = (\mathcal{V}_u^M, \mathcal{E}_u^M)$. In this setup, $\mathbf{A} \in \mathbb{R}^{|\mathcal{V}_u^M| \times |\mathcal{V}_u^M|}$ represents the adjacency matrix of the molecular graph, and $\mathbf{X}^{(0)} \in \mathbb{R}^{|\mathcal{V}_u^M| \times 107}$ is the initial node feature matrix, where each row encodes a 107-dimensional feature vector for an atom. Specifically, the first 92 dimensions are derived using the CGCNN \cite{PhysRevLett_CGCNN} framework, capturing fundamental atomic characteristics such as atomic number, atomic radius, and electronegativity. The remaining 15 dimensions are aggregated features of all chemical bonds connected to the atom, encoding properties such as bond type, directionality, length, and ring status (detailed in Table \ref{tab:atom_bond_features}) \cite{li2024kagnn}. The GCN iteratively propagates and aggregates these features across the molecular graph in the following rule:
\begin{equation}\label{GCN}
    \mathbf{X}^{(l+1)} = \sigma\left( \tilde{\mathbf{D}}^{-\frac{1}{2}} \tilde{\mathbf{A}} \tilde{\mathbf{D}}^{-\frac{1}{2}} \mathbf{X}^{(l)} \mathbf{W}_M^{(l)} \right),
\end{equation} 
where $\tilde{\mathbf{A}} = \mathbf{A} + \mathbf{I}_{|\mathcal{V}_u^M|}$ is the adjacency matrix with added self-connections, $\tilde{\mathbf{D}}$ is the diagonal degree matrix of $\tilde{\mathbf{A}}$, $\mathbf{W}_M^{(l)}$ denotes the trainable weight matrix for the $l$-th layer, and $\sigma$ represents the ReLU activation function. By stacking $L_4$ such layers, the GCN refines the node embeddings, $\mathbf{X}^{(L_4)}$, effectively capturing both local structural information and chemical properties. To obtain a unified representation of the entire molecular graph, we apply a SumPooling operation across all node embeddings:
\begin{equation}
    \mathbf{h}_u^M = \mathrm{SumPooling} \left(\mathbf{X}^{(L_4)} \right).
\end{equation}
The resulting vector $\mathbf{h}_u^M$ encodes the drug molecule's structural and chemical properties, enabling practical downstream analysis.

\begin{table*}[ht]
\centering
\caption{List of atom and bond features.}
\label{tab:atom_bond_features}
\renewcommand{\arraystretch}{1.2} 
\setlength{\tabcolsep}{10pt} 
\begin{tabular}{llp{7cm}cl}
\toprule
\textbf{Feature Type} & \textbf{Feature Name} & \textbf{Description} & \textbf{Size} & \textbf{Type} \\
\midrule
\multirow{1}{*}{\textbf{Atom Feature}} 
                      & CGCNN & Atomic number, Radius, and electronegativity. & 92 & One-hot \\
\midrule
\multirow{4}{*}{\textbf{Bond Feature}} 
                      & Bond Directionality &  None, Beginwedge, Begindash, etc. & 7 & One-hot \\ 
                      & Bond Type & Single, Double, Triple, or Aromatic. & 4 & One-hot \\
                      & Bond Length & Numerical and square length of the bond. & 2 & Float \\
                      & In Ring & Whether the bond is part of a chemical ring. & 2 & One-hot \\ 
\bottomrule
\end{tabular}
\end{table*}

\begin{table*}[ht]
\centering
\caption{Effect of InfoMin on Generalization Performance. }
\begin{tabular}{llcccc}
\toprule
\textbf{Dataset} & \textbf{Model} & \textbf{Train F1} & \textbf{Test F1} & \textbf{Gen. Error} (Train F1-Test F1) & \textbf{Gen. Ratio} (Test F1/Train F1) \\
\midrule
\multirow{2}{*}{Ryu} & MolecBioNet & 0.9999$\pm$0.0001 & 0.9124$\pm$0.0079 & 0.0874$\pm$0.0080 & 0.9126$\pm$0.0080 \\
                     & w/o InfoMin & 0.9985$\pm$0.0020 & 0.9003$\pm$0.0136 & 0.0982$\pm$0.0126 & 0.9016$\pm$0.0127 \\
\midrule
\multirow{2}{*}{DrugBank} & MolecBioNet & 0.9986$\pm$0.0018 & 0.8465$\pm$0.0098 & 0.1521$\pm$0.0092 & 0.8477$\pm$0.0093 \\
                          & w/o InfoMin & 0.9963$\pm$0.0030 & 0.8068$\pm$0.0086 & 0.1895$\pm$0.0060 & 0.8098$\pm$0.0065 \\
\bottomrule
\end{tabular}
\label{tab: generalization ability of infomin}
\end{table*}

\section{Sensitivity Analysis}

\begin{figure}[h]
  \centering

  \subfigure{
    \includegraphics[width=0.475\linewidth]{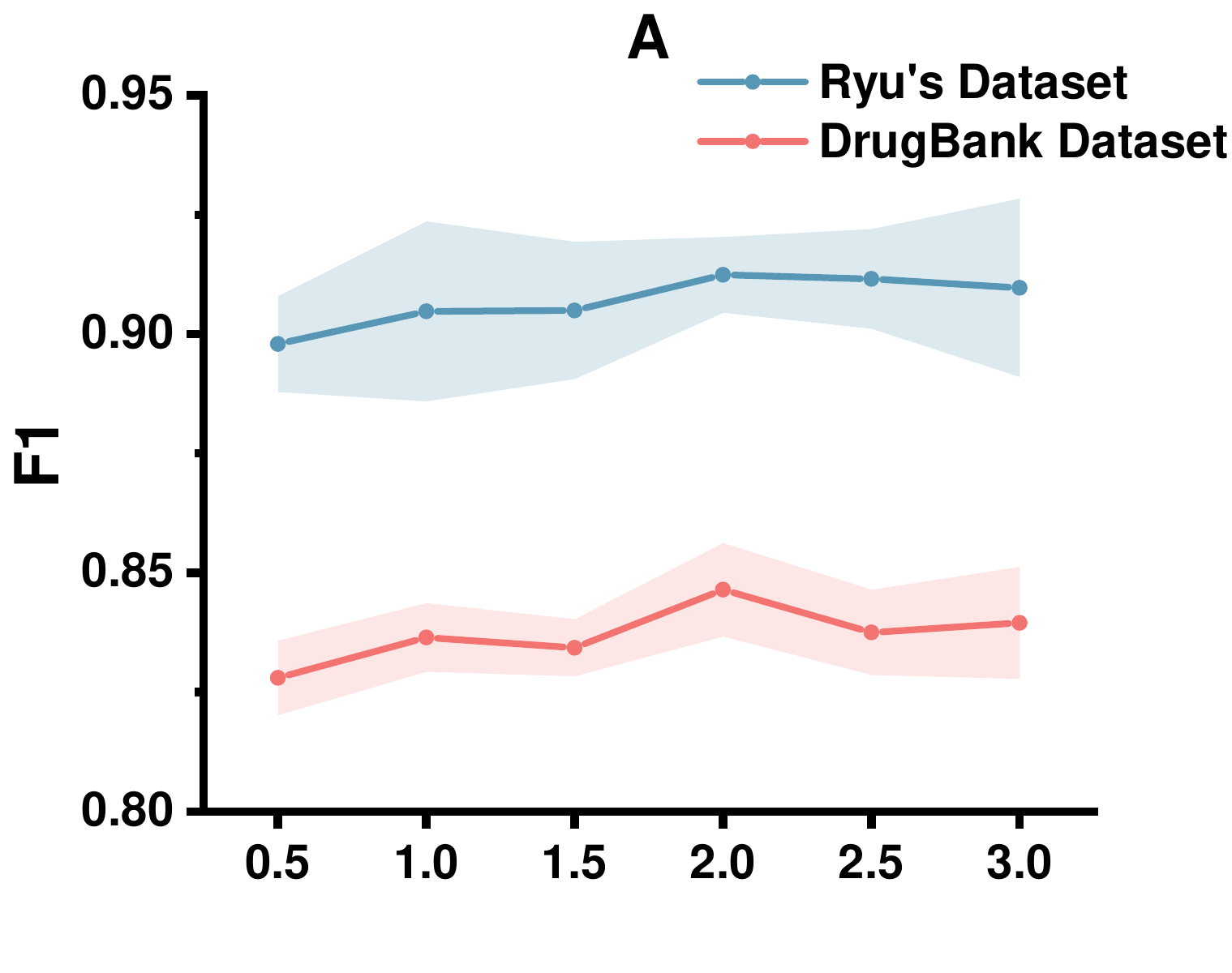}
    \label{fig:Sensitivity experiment_beta}
  }
  \hfill
  \subfigure{
    \includegraphics[width=0.475\linewidth]{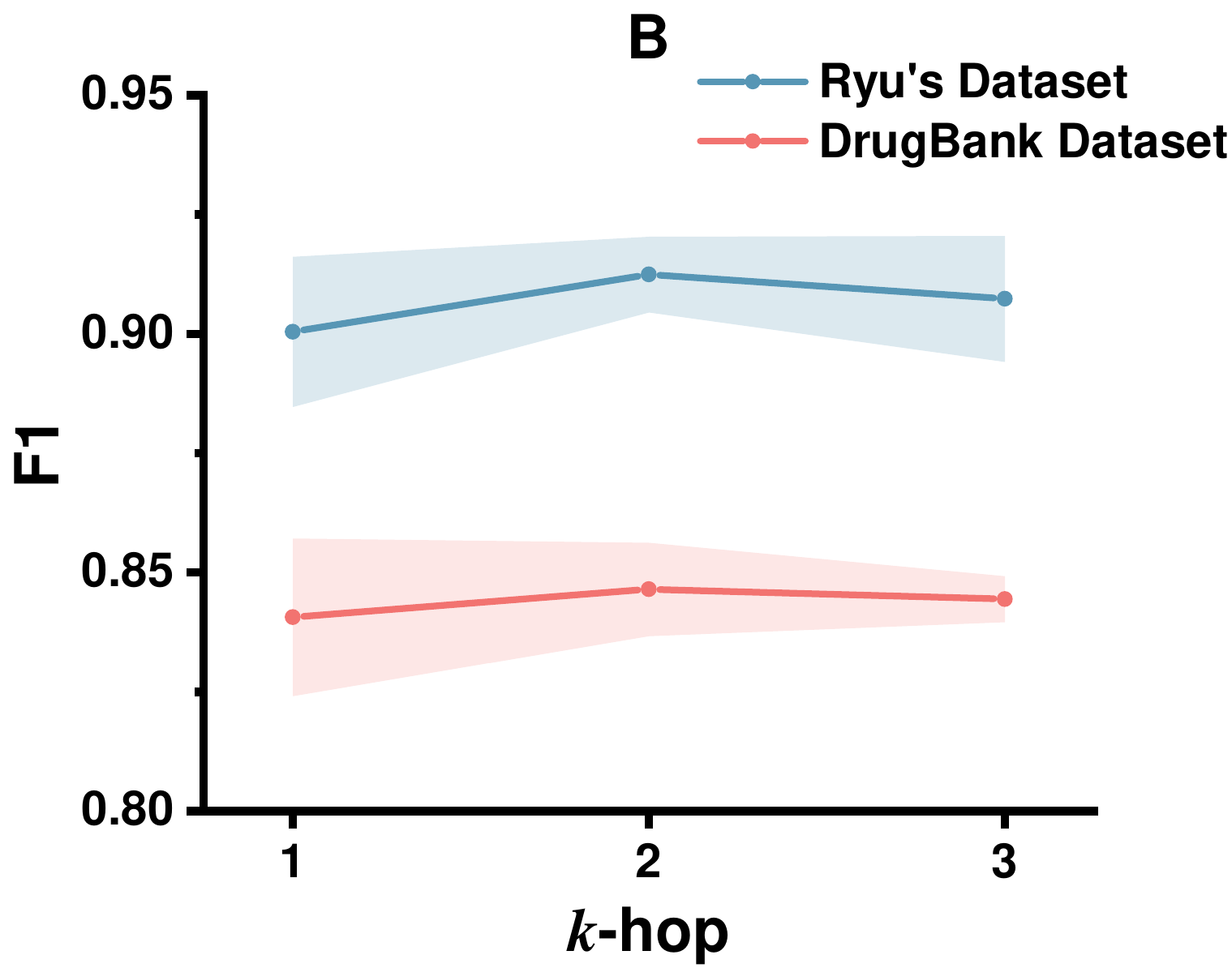}
    \label{fig:Sensitivity experiment_k}
  }

  \caption{Sensitivity analysis of hyperparameters on model performance. (A) The effect of $\beta$, the weighting factor for the center loss $\mathcal{L}_C$, and (B) the influence of $k$, which determines the size of the $k$-hop enclosing subgraph for each drug pair. Results are reported as mean F1 scores with standard deviation across five-fold cross-validation.}
  \label{fig: Sensitivity experiment}
\end{figure}

We analyze the effect of key hyperparameters in our framework, with the results presented in Fig.~\ref{fig: Sensitivity experiment}. In particular, Fig.~\ref{fig: Sensitivity experiment}A shows the performance trend as the center loss weight $\beta$ varies from 0.5 to 3. We observe that the F1 score generally increases initially, indicating that incorporating center loss is beneficial. While the model's performance stabilizes on the Ryu dataset and fluctuates slightly on the DrugBank dataset, both achieve peak performance at $\beta = 2$. Therefore, we adopt $\beta = 2$ as the default setting. As for the mutual information loss coefficient $\gamma$, we set its value by observing that the prediction loss is approximately ten times larger than the mutual information loss, and scale $\gamma$ accordingly to maintain a balanced optimization.

We also investigate the effect of the $k$ parameter in $k$-hop enclosing subgraphs. As shown in Fig.~\ref{fig: Sensitivity experiment}B, model performance improves from $k=1$ to $k=2$, but slightly declines at $k=3$. We default to $k=2$ in MolecBioNet, as it strikes a good balance between capturing meaningful biomedical context and avoiding unnecessary complexity. Two-hop neighborhoods already include the main types of biomedical entities relevant to DDIs, such as proteins, pathways, diseases, side effects, chemical compounds, and genetic disorders, while extending to 3-hop brings little additional diversity or relevance. Moreover, increasing $k$ to 3 substantially enlarges the subgraphs, especially in densely connected biomedical graphs, leading to higher computational cost and potential dilution of informative signals. These findings support $k=2$ as an effective and efficient choice.

\section{Additional Results and Visualizations}

\begin{figure}[ht]
  \centering
  \includegraphics[width=\linewidth]{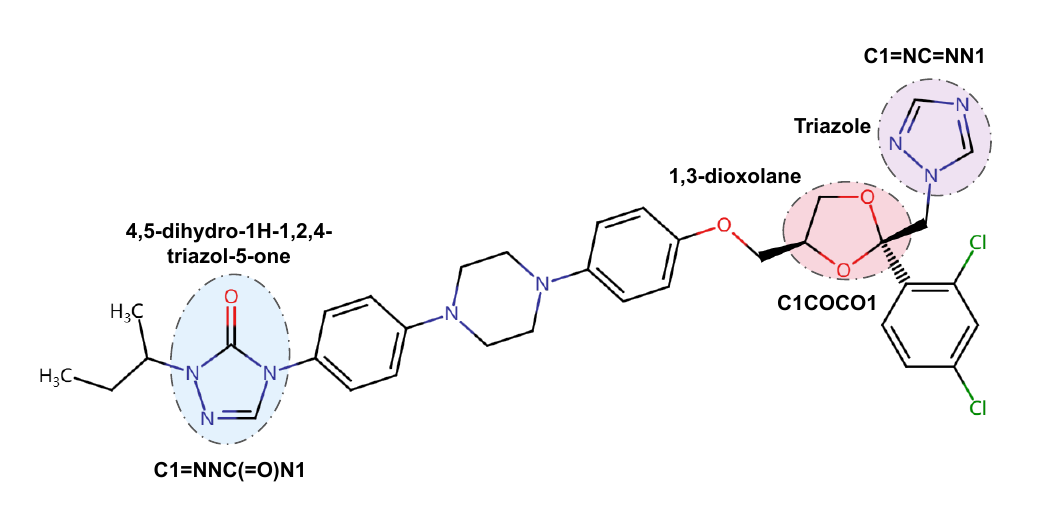}
  \caption {Chemical substructures in Itraconazole that influence the metabolism of Promazine. The top three substructures are identified through attention-guided influence pooling, with Triazole showing the greatest impact by binding to the active site of CYP3A4.}
  \label{fig: Itraconazole.}
\end{figure}

\begin{figure}[ht]
  \centering
  \includegraphics[width=\linewidth]{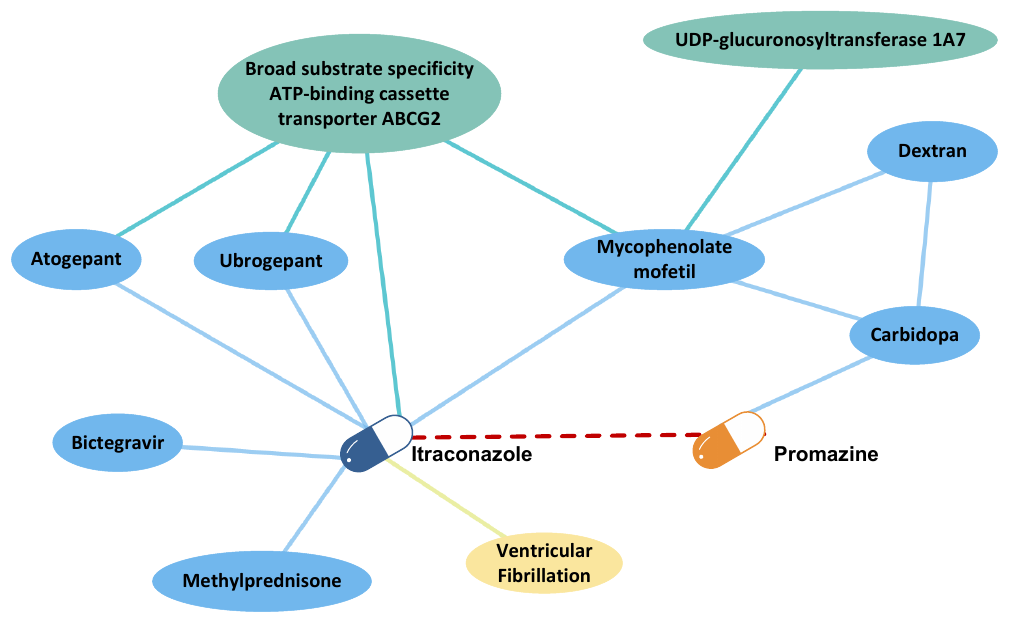}
  \caption {Local subgraph extracted from tsBKG for Itraconazole and Promazine, showing the top 10\% of most influential nodes.}
  \label{fig: explain.}
\end{figure}

This section includes supplementary generalization results (Table~\ref{tab: generalization ability of infomin}) and interpretability visualizations (Fig.~\ref{fig: Itraconazole.} and Fig.~\ref{fig: explain.}) to support the main analysis.

\end{document}